# Development, Evaluation, and Deployment of a Multi-Agent System for Thoracic Tumor Board

Tim Ellis-Caleo, MD[1*]; Timothy Keyes, PhD[2,3*]; Nerissa Ambers, MPH[4]; Faraah Bekheet, MD[5]; Wen-wai Yim, PhD[6]; Nikesh Kotecha, PhD[2]; Nigam H. Shah, MBBS, PhD[2,3]; Joel Neal, MD, PhD[1,7]

1. Division of Oncology, Department of Medicine, Stanford University School of Medicine, Palo Alto, CA
2. Technology and Digital Solutions, Stanford Health Care, Palo Alto, CA
3. Department of Biomedical Data Science, Stanford University School of Medicine, Palo Alto, CA
4. Nursing Informatics, Stanford Health Care, Palo Alto, CA
5. Department of Medicine, Stanford University School of Medicine, Palo Alto, CA
6. Microsoft AI, Redmond, WA
7. Stanford Cancer Institute, Palo Alto, CA

# Abstract

Tumor boards are multidisciplinary conferences dedicated to producing actionable patient care recommendations with live review of primary radiology and pathology data. Succinct patient case summaries are needed to drive efficient and accurate case discussions. We developed a manual AI-based workflow to generate patient summaries to display live at the Stanford Thoracic Tumor board. To improve on this manually intensive process, we developed several automated AI chart summarization methods and evaluated them against physician gold standard summaries and fact-based scoring rubrics. We report these comparative evaluations as well as our deployment of the final state automated AI chart summarization tool along with post-deployment monitoring. We also validate the use of an LLM as a judge evaluation strategy for fact-based scoring. This work is an example of integrating AI-based workflows into routine clinical practice.

# Introduction

Large language models (LLMs) have rapidly expanded the scope of tasks that can be performed using unstructured clinical text.[1,2] These tasks are particularly relevant in oncology, where clinical decisions often depend on integrating heterogeneous, longitudinal data, often in textual form, that are fragmented throughout a patient's electronic health record (EHR). Tumor boards—multidisciplinary patient care conferences in which clinicians review patient-specific information to formulate consensus recommendations for diagnosis and treatment—depend on rapid, accurate presentation of a patient's longitudinal clinical history, imaging, pathology, and genomic findings to support shared reasoning. In thoracic oncology, tumor boards have been associated with improved care processes and patient outcomes, yet their effectiveness can be limited by the burden of synthesizing longitudinal case histories for complex patients and by competing time pressures on clinicians.[3] These constraints can limit the adoption and effectiveness of tumor boards across institutions and require approaches that reduce documentation burden while improving information-gathering effectiveness.

Prior work has evaluated whether LLMs can generate tumor board-style treatment recommendations, reporting moderate-to-high concordance with expert consensus in retrospective thoracic, gastrointestinal, and breast cancer cohorts[4–7]. More recently, agentic architectures—in which an LLM plans and executes multistep actions using tools[8]—have been proposed to better approximate the reasoning required for tumor board decision support[9]. However, these studies have generally focused on whether AI systems can replicate or replace tumor board outputs, often using retrospectively curated inputs or comparing autonomous agents against LLM baselines that lack access to the same underlying data—conflating performance gains from autonomy with those from data access. This framing overlooks a more immediate operational need: high-functioning tumor boards already produce sound recommendations, but clinicians bear a substantial burden in synthesizing longitudinal case histories to prepare for them. It also leaves health systems without empirical guidance about when higher-autonomy agentic designs (e.g. tool-using or multi-agent systems) outperform simpler, constrained workflows—or when their added complexity primarily introduces new failure modes and cost[10,11].

In this work, we addressed a pragmatic operational and quality improvement need within Stanford Health Care's (SHC) thoracic tumor board: generating concise, tumor board-ready patient summaries from longitudinal EHR documentation with a focus on preserving clinically relevant details critical for decision making. To do so, we performed a comparative evaluation of five automated LLM-based summarization systems spanning a spectrum of complexity and autonomy—from constrained, single-step LLM workflows to tool-enabled, multi-agent architectures—using two summaries (a physician-authored reference summary and a manually-generated LLM summary used during routine operations) as comparators. Using objective rubric-based fact scoring and subjective physician ratings of usability, relevance, and accuracy, we identified the LLM system most suitable for deployment in routine practice. We then integrated the selected system into the tumor board workflow and conducted post-deployment quality monitoring to characterize ongoing performance and to detect potential patient-safety–relevant degradation beyond the initial evaluation set. By pairing a comparative evaluation across systems with differing levels of complexity with a monitored, prospective deployment, we provide empirical guidance on performance and when added autonomy affects tumor-board summarization under real operational constraints.

# Methods

## Stanford Thoracic Tumor Board

The Thoracic Tumor board at Stanford Health Care is a weekly, hour-long meeting attended by medical oncology, radiology, thoracic surgery, pathology, radiation oncology, and pulmonology. Typically, 12-20 cases are reviewed each week based on clinician submission and are selected when multidisciplinary and/or colleague input is needed for complex, time-sensitive decisions. For each case, radiology and pathology images are reviewed with live interpretation by domain specialists, followed by multidisciplinary discussion and a resulting actionable recommendation documented in the electronic health record (EHR) by the presenting clinician.

Because tumor boards are non-billable, and there is significant opportunity cost for the attendee's time, limited time is preferable per case (approximately 3-5 minutes on average).  To make discussion more efficient, it is grounded by an orally presented longitudinal summary of a patient's clinical course, consistent with the "summary statement" format often used in oral case presentations to support shared reasoning[12–14].

## Manual LLM (SecureGPT) Workflow Integration

As a quality improvement intervention, we integrated an AI-assisted chart summarization workflow into the thoracic tumor board process. Clinic administrative staff used SecureGPT (SHC's HIPAA-compliant LLM environment[15]) to generate a brief tumor board summary by prompting the model with a thoracic oncology-specific prompt and the patient's most recent relevant (per their discretion) clinical progress note.  The resulting summary was constrained to 999 characters (for pasting into virtual meeting chat boxes), saved as a Smart Data Element, and displayed in the tumor board visit whiteboard (an Epic

Smart Form) in the EHR. During tumor board, the summary was displayed alongside radiology and pathology review, using the meeting chat feature, to ground discussion in key clinical details.

Although feasible in routine practice, this human-in-the-loop process did not scale well. It depended on manual selection of a single clinical note, required repeated staff time for each case, and was subject to copy-and-paste error. With the limitations of the manual workflow in mind, we started a quality improvement intervention to design, test, and deploy an automated LLM-based workflow for tumor board chart summarization.

## Intervention Design and Outcomes

We conducted a retrospective comparative evaluation of tumor board case summaries generated by five different automated, LLM-based summary methods and compared them to two baseline comparators—a physician-authored reference summary and the manually generated SecureGPT summary used operationally prior to the quality improvement project. The evaluation was designed to support selection of a summary-generation approach suitable for routine deployment.

We prespecified two primary evaluation axes:

1. **Objective summary completeness**, measured as rubric-based fact-scoring assessed using a validated LLM-as-a-judge pipeline. This endpoint was chosen to quantify clinically relevant omissions and inaccuracies, motivated by prior work suggesting that omissions are a prominent concern in LLM-generated clinical summaries[16,17].

2. **Subjective summary quality**, measured by physician ratings across four domains adapted from prior work[18]: Overall, Writing Style, Relevance, and Accuracy, using structured rubrics where scores ≥3 (on a 1–5 scale) were prespecified as acceptable for tumor board use.

In addition, to assess whether retrospective findings were consistent with routine clinical operation, we performed post-deployment quality monitoring using subjective physician ratings along the same axes described above.

## Patient Cohort

We included a randomly chosen cohort of thoracic tumor board cases from April through June 2025 (n=50), resulting in a representative cohort of cases discussed during a single academic quarter.

# Summary Generation

## Physician-Authored Reference Summaries

For each patient, a board-certified medical oncologist authored a concise tumor board–style summary after comprehensive chart review. The oncologist was blinded to all model-generated summaries and was instructed to simulate routine tumor board preparation, producing a short discussion-oriented summary suitable for viewing during the live presentation.

## SecureGPT Manual Workflow Summaries

During routine clinical care prior to the quality improvement project, a summary was created manually using SecureGPT and used in the thoracic tumor board. These were retrieved and scored against the reference summaries as a baseline.

## Automated Summary Generation Methods

Following initial adoption of the manually-generated SecureGPT summaries, we implemented five automated systems to generate tumor board–ready summaries without manual prompting. All automated methods were instructed to produce a succinct summary tailored to tumor board use with standardized headings ("ID," "Biomarkers/Next-Generation Sequencing (NGS)," and "Prior therapy") and a 999-character limit to mirror operational constraints. Automated systems used GPT-4.1 (OpenAI) as the underlying LLM, with deterministic decoding (temperature 0) and were accessed via ChatEHR, SHC's proprietary platform for LLM integration with the EHR.[19]  All automated systems drew from clinical notes as stored in the EHR without preprocessing (no template stripping, deduplication, or redaction). Methods that incorporated longitudinal context were restricted to clinical notes within a fixed 180-day lookback window preceding the tumor board date unless stated otherwise. Prompts for all systems are provided in **Appendix 1**.

*Automated summary-generation approaches*

We evaluated five automated summary-generation approaches:

1. **Single-note (most recent oncology note).** The system summarized only the patient's most recent oncology note prior to the scheduled tumor board date. This method most closely paralleled the manual SecureGPT workflow.

2. **Single-step concatenation.** The system concatenated all notes from the 180-day window preceding the tumor board date in chronological order and produced a summary in one LLM call.

3. **Multi-step summarization.** The system first generated constrained per-note extracts (≤5 lines) capturing note date and key tumor board fields and then performed a final synthesis step to produce the final tumor board summary.

4. **Low-autonomy multi-agent system.** A multi-agent system retrieved notes once from the fixed 180-day window and followed detailed, cancer-specific instructions specifying which notes and facts were relevant and what content to include in each section of the final summary.

5. **High-autonomy multi-agent system.** A multi-agent system could adaptively select the retrieval time window on a per-patient basis, call the note-retrieval tool iteratively if initial retrieval was insufficient, and used only high-level guidance to decide which notes and clinical details were relevant for tumor board discussion. It retained the same three-section organization for the final summary but had flexibility in how it structured content within each section.

*Operational definition of autonomy*

In this study, we define "autonomy" as the extent to which an LLM system dynamically determined data retrieval scope, tool use, and note selection rather than executing a fully specified procedure. Compared to the low-autonomy multi-agent system, the high-autonomy multi-agent system differed in three ways: (1) its data retrieval window selection was adaptive (could be chosen by the LLM) rather than fixed; (2) data retrieval could be repeated iteratively rather than performed a single time; and (3) note selection and content inclusion were guided by higher-level instructions rather than comprehensive, thoracic cancer-specific directives (see **Appendix 1** for full prompt details).

*Model configuration and orchestration*

Single-note, single-step, and multi-step pipelines were orchestrated in LangChain[20]. Agentic pipelines were orchestrated using the Microsoft Healthcare Agent Orchestrator (HAO)[21]. Outputs were persisted to secure institutional storage for analysis. Full prompts and system specifications are provided in **Appendix 1**.

## Development of a Fact-Based Scoring Rubrics

For each patient, we created a structured rubric enumerating key facts expected in an optimal tumor board summary. Rubrics were organized as high-level facts ("attributes") that could be optionally decomposed into constituent details ("subattributes") to support more granular scoring. Each attribute/subattribute was labeled by information type: Demographics, Stage, Pathology, Molecular Findings, Medical Treatment, Surgical Treatment, or Radiation Treatment. Typically, parent attributes captured a clinically salient item (e.g., "Resection in 2021"), and subattributes captured constituent details such as procedure type ("Resection") and timing ("in 2021").

We evaluated rubric items at the highest available level of granularity: attributes without subattributes were treated as a single rubric item, whereas for attributes with subattributes, only subattributes were included (the parent attribute was not redundantly scored in the analysis) to avoid double-counting.

Rubrics were authored following detailed review of the medical record by a single board-certified medical oncologist who routinely attends the thoracic tumor board and was familiar with the patient population. Facts were selected based on clinical judgment emphasizing items likely to influence multidisciplinary management. Across patients, the median number of attributes was 6 (IQR 5-7) and the median number of subattributes was 3.5 (IQR 2-6); ranges were 2-12 for attributes and 0-16 for subattributes.

## Evaluation of Rubric-based Fact-scoring using an LLM-as-Judge

We developed an LLM-as-a-judge pipeline to assess fact scoring for each patient-summary pair based on previously published work[22]. Rubrics were serialized into structured Pydantic objects[23] (attributes with nested subattributes), and a standardized evaluation prompt instructed the model to assign scoring labels for each attribute and subattribute: *Yes* (present in the summary), *Partial* (partially present in the summary), or *No* (absent from the summary). When the assigned scoring label was not *Yes*, the judge also assigned an error type (Missing, Incorrect, Ambiguous, or Other). The prompt permitted semantic equivalence without requiring exact phrasing. Consistent with the primary analysis strategy, rubric items were analyzed at the highest available granularity (attributes without subattributes, and subattributes otherwise).

The judge model was GPT-5 (OpenAI) accessed via a LiteLLM proxy with credentials managed through Azure Key Vault. For all judge runs, reasoning effort was set to "medium," with other parameters at defaults.

To validate the LLM-as-a-judge's fact-scoring against physician assessment, two physicians independently scored a pilot set of summaries (n = 250 summaries) using the same rubrics and labels (*Yes*/*Partial*/*No*), labeling each attribute/subattribute as present, partially present, or absent. These physician labels were used solely to quantify agreement with the LLM judge and to support use of the LLM judge for scalable scoring evaluation in the main comparative study. Judge prompts are provided in **Appendix 2**.

## Subjective Ratings by Physicians

Two physicians independently evaluated each summary using four structured rubrics: Overall, Writing Style, Relevance, and Accuracy, each on a 1-5 scale (**Appendix 3**). Rubrics were designed such that ratings ≥3 were prespecified as acceptable for tumor board use. For each domain, the final score was defined as the average of the two physician ratings; when the mean fell between Likert categories, it was rounded down to the lower category to apply a more conservative threshold for acceptability.

## Post-Deployment Quality Monitoring

Automated summaries were introduced into SHC's thoracic tumor board workflow beginning in January 2026. To assess whether summary quality remained acceptable outside the retrospective evaluation set, a board-certified medical oncologist who attended every tumor board meeting and an Internal Medicine resident independently rated 50 production summaries from the first month deployment using the same structured rubrics described above. These were the same two physicians who performed the pre-deployment analysis. As above, the final score in each domain was defined as the average of the two ratings and, when the mean fell between Likert categories, was rounded down to the lower category. These results were analyzed descriptively as a post-deployment quality signal.

## Statistical Analysis

Analyses were conducted in Python[24], with statistical analyses and visualization performed in R using the tidyverse ecosystem[25]. For objective summary completeness (fact scoring analysis), we computed summary-level rubric-based fact-scoring metrics derived from the LLM judge, defined as the proportion of rubric items present within each summary. We reported two complementary fact-scoring definitions: (1) "fully present", in which only Yes was scored as a success and (2) "fully or partially present" in which both Yes and Partial were scored as successes. Metrics were summarized overall and stratified by attribute/subattribute type. For comparisons across automated methods, we used within-case nonparametric tests (Friedman rank sum test) and prespecified paired post-hoc comparisons (Wilcoxon signed-rank tests) between the current-state (SecureGPT) and each automated LLM system[26]. To adjust for multiple comparisons, we used false-discovery rate (Benjamini-Hochberg) adjustment[27].

For subjective summary quality, physician ratings were treated as ordinal outcomes and compared within-case using paired Wilcoxon signed-rank tests[26]. We contrasted the current-state workflow (SecureGPT) with each automated system separately for each rating domain (Overall, Writing Style, Relevance, Accuracy). P-values were adjusted for multiple comparisons using false-discovery rate (Benjamini-Hochberg) adjustment [27].

For validation of the LLM judge, we quantified agreement between the judge and physician fact-scoring labels using exact agreement and chance-corrected measures (Cohen's kappa and Gwet's AC1) on the preliminary summaries, with 95% confidence intervals obtained via bootstrap resampling clustered at the patient level. We also assessed multi-rater agreement using Fleiss' kappa as previously described[28].

# Results

We evaluated tumor board-oriented patient summaries for 50 thoracic tumor board cases presented from April through June 2025. For each case, we assembled seven summaries: a physician-authored reference summary, the manually generated SecureGPT summary used operationally during tumor board, and five automated summaries spanning a spectrum of autonomy and system complexity. All automated systems were constrained to the same output format (999 characters; standardized headings) to support comparability across generation approaches.

## Validation of the LLM-judge against physician fact-scoring labels

Before comparing summary-generation methods, we evaluated whether the LLM-as-a-judge pipeline produced scoring labels that were concordant with physician assessment. Two physicians independently scored a pilot set of patient-summary pairs (50 patients × 5 pilot summary generation methods = 250 summaries) using the same case-specific rubrics and the same three-level scoring labels (Yes/Partial/No) applied in the primary analyses. Agreement between the judge and physicians was assessed at the rubric-item level, using exact agreement as well as chance-corrected measures (Cohen's kappa and Gwet's AC1), with 95% confidence intervals estimated by bootstrap resampling clustered at the patient level (**Supplemental Figures 1-6**). Across rubric items, the judge demonstrated high concordance with physician labels (exact agreement = 89.3-89.4%; Cohen's κ = 0.687-0.709; Gwet's AC1 = 0.870-0.871), supporting use of the judge for scalable fact-based scoring in the full comparative evaluation.

We also examined multi-rater consistency among the three raters (two physicians and the judge) using Fleiss' κ to contextualize judge-physician agreement relative to overall inter-rater reliability (**Supplemental Figure 6**). Fleiss' κ was 0.731 (95% CI 0.697-0.762) between only the 2 physicians and 0.708 (95% CI 0.684-0.734) when the LLM was included as an additional rater.

Because the primary analyses summarize fact-based scoring at the patient level, we also compared patient-level completeness metrics computed from physician and judge fact scoring labels (proportion fully present in **Supplemental Figure 7**; proportion fully or partially present in **Supplemental Figure 8**), which were strongly correlated across cases for both definitions. Taken together, these findings suggest that the LLM judge's scoring labels track clinician judgments sufficiently to support comparative analyses of omission- and error-prone domains across automated summarization methods.

## Objective summary completeness (rubric-based fact-scoring)

Across the cohort, rubric-based fact-scoring differed by generation approach for both fully (Friedman rank-sum $\chi^2(5)$ = 57.0; $p < 0.001$) and partially ($\chi^2(5)$ = 50.7; $p < 0.001$) present facts (**Figure 1**). The single-note baseline—summarizing only the most recent oncology progress note—showed the lowest completeness and did not significantly improve fact scoring relative to the current-state manual SecureGPT workflow for fully (Wilcoxon signed-rank test V = 474, p = 0.169) or partially present facts (V = 398, p = 0.052), consistent with the observation that key tumor-board facts were often distributed across multiple notes in the longitudinal record.

In contrast, expanding the input scope to a fixed longitudinal window improved completeness. Both the single-step method (summarizing all notes from the prior 180 days in a single LLM call) and the multi-step method (per-note extraction followed by a synthesis step) achieved significant increases in fact-scoring compared to the SecureGPT baseline (fully present V = 16.5 and V = 17.5, respectively; both p < 0.001; partially present V = 18 and V = 23.5, respectively; both p < 0.001).

Results for the multi-agent systems differed based on the degree of autonomy afforded to each system. The low-autonomy multi-agent system—which was specifically instructed to use a fixed 180 day data retrieval window and to apply strict criteria for note relevance and content inclusion in the final summaries—also demonstrated a significant increase in fact-scoring compared to the SecureGPT baseline (fully present V = 40.5, p < 0.001; partially present scoring  V = 50.0, p = 0.001). Notably, increased autonomy did not confer a completeness advantage: unlike the low-autonomy multi-agent system, the high-autonomy multi-agent system did not significantly improve scoring relative to SecureGPT (fully present  V = 145, p = 0.221; partially present V = 164, p = 0.380).

When rubric-based fact-scoring was stratified by attribute/subattribute type (Demographics, Stage, Molecular, Pathology, Medical Treatment, Radiation Treatment, and Surgical Treatment), rates differed across domains (**Table 1**). Simple demographic attributes were captured at high rates across summary generation methods. However, both the manual SecureGPT workflow and the single-note system showed substantially lower completeness for staging and molecular findings—consistent with these details being distributed across multiple documents in a patient's record and often requiring synthesis of multiple data points to assemble. The single-note system performed worst for longitudinal domains, with marked reductions in medical and radiation treatment fact-scoring relative to SecureGPT, underscoring that a single oncology note frequently fails to reflect the patient's cumulative treatment history due to either differential focus of the note or omission.

By contrast, methods that incorporated a larger longitudinal lookback window (single-step, multi-step, and low-autonomy multi-agent) improved completeness across multiple domains, with especially large gains for medical, radiation, and surgical treatment history attributes. Notably, for Biomarkers/NGS, improvements were more pronounced for full-or-partial fact presence than for full fact presence, suggesting that automated summaries often captured biomarker information incompletely. Finally, the high-autonomy agentic system was less likely to correctly include facts about pathology and radiation treatment, suggesting it may have struggled to identify details from pathology reports and radiation oncology notes as relevant for the tumor board summary compared to LLM systems constrained by more specific instructions.

Interestingly, cancer stage was a consistent outlier: it was captured almost universally in physician-authored summaries yet remained substantially less complete in every automated method, suggesting that cancer staging is a persistent abstraction challenge for LLM systems even when longitudinal context is provided.

## Subjective physician ratings of usability and quality of summaries

Across all paired ratings, agreement between the two physician reviewers was moderate: exact agreement occurred in 62.4% of ratings, and reviewer scores were moderately correlated overall (Spearman ρ = 0.669). Differences between reviewers were centered near zero, with most discordant ratings (77.4% of discordant ratings) differing by 1 Likert point rather than reflecting large disagreements (**Supplemental Figure 9**).

Physician ratings across the four subjective domains—overall usefulness, writing style, relevance, and accuracy—revealed important tradeoffs that were not fully captured by rubric-based fact-scoring alone (**Figure 2**). The single-note baseline was consistently rated worse than the current-state manual SecureGPT workflow across domains, with significantly lower ratings for overall (Wilcoxon signed-rank test V = 602, p = 0.001), relevance (V = 724, p < 0.001), writing style (V = 620, p = 0.005), and accuracy (V = 410, p = 0.005).

In contrast, methods that incorporated a wider longitudinal window (single-step, multi-step, and both the low- and high-autonomy multi-agent systems) did not show statistically significant improvements over SecureGPT on subjective measures, despite their higher fact-scoring. Notably, both single-step (V = 591, p = 0.005) and multi-step (V = 354, p = 0.024) approaches were rated significantly lower on relevance compared to SecureGPT, suggesting that naively expanding context may increase inclusion of extraneous information and reduce the salience of decision-critical details for clinicians under time pressure (**Figure 2**). Consistent with this, the single-step approach also received significantly lower overall ratings than SecureGPT (V = 496, p = 0.022), whereas the multi-step approach did not (V = 232, p = 0.243). No significant differences were detected between SecureGPT and either multi-agent system in any subjective domain.

## Post-deployment monitoring of subjective summary quality

On the basis of the pre-deployment comparative evaluation, we selected the low-autonomy multi-agent system for production deployment. This system met the prespecified deployment criterion of improving objective completeness relative to the current-state manual SecureGPT workflow in fact-scoring analyses, while avoiding the reduction in clinician-perceived relevance observed with the longitudinal single-step and multi-step approaches. A schematic of the deployed workflow and its agentic control logic is shown in **Figure 3**.

The deployed multi-agent system comprised four specialized agents called sequentially: *Data Loader Agent*, *FHIR (Fast Healthcare Interoperability Resource) Agent*, *Curation Agent*, and *Summarization Agent* (**Figure 3**). The *Data Loader Agent* delegated EHR retrieval to the *FHIR Agent* via an inter-agent handoff. The *FHIR Agent* invoked a single EHR data-access capability parameterized by patient identifier, time window, and resource type to retrieve clinical notes, then wrote the results to an ephemeral workspace as timestamped text files. The *Curation Agent* then applied a workspace filter that evaluated each note's tumor-board relevance and removed nonrelevant items in place with an auditable rationale. Finally, the *Summarization Agent* synthesized the final summary using only information explicitly documented in the chart and persisted the artifact together with per-statement citations. During persistence, the storage capability resolved cited notes by identifier, computed character offsets for quoted snippets within the source text, and saved the summary to durable storage under a workflow-instance identifier.

Following the deployment of this system, the same board-certified medical oncologist and Internal Medicine resident rated 50 deployed summaries across the same four subjective domains evaluated before deployment (Overall, Writing Style, Accuracy, and Relevance)

(**Figure 4**). During the first post-deployment month, all rated summary ratings remained above the prespecified minimum acceptable threshold for use in tumor board (score ≥3).

## Discussion

We developed an automated LLM-based chart summarization tool and deployed it in routine clinical practice at the SHC thoracic tumor board as a quality improvement initiative. In contrast to prior work, our process did not seek to replace tumor boards but instead augment them. We found a clear divergence between the objective, fact-based scoring evaluation of our automated summaries when compared to the subjective axes. Rubric-based fact-scoring captured clinically important differences that were not necessarily captured in subjective physician ratings. Although several of the LLM systems we evaluated (single-step, multi-step, and low-autonomy multi-agent systems) produced more complete summaries (with a higher proportion of fully present key facts), clinicians did not consistently rate these summaries as better overall. From an implementation perspective, this supports using rubric-based completeness as a complementary, safety-oriented endpoint—particularly when the goal is to avoid silent loss of key facts that can dramatically change medical decision making—rather than relying solely on global subjective assessments of usefulness or style.

Attribute-level analyses informed two key areas where the automated systems performed poorly compared to physician authored summaries—next generation sequencing and cancer staging. Across automated methods, molecular/Next Generation Sequencing (NGS) information was occasionally missing despite nearly always being included in physician summaries. In our environment, this reflects an important technological limitation: key genomic results are often stored outside standard note text (e.g., in the EHR's media tab or attached reports), which were not accessible to SecureGPT or at the time of the study. This has practical implications for roadmap planning: expanding tool access to these data sources is likely a higher-yield intervention for improving NGS completeness than further prompt or architecture iteration alone. It also motivates more intensive data structuring the in EHR to fully integrate third-party testing results as discrete data. Despite these limitations, we found this to be a highly clinically useful tool for everyday practice.

Cancer stage was nearly universal in physician-authored summaries but remained substantially less complete across all automated methods—even when longitudinal context was provided. This is likely because cancer staging is a complex abstraction problem requiring reconciliation of evolving pathology, imaging, and clinical assessments over time. Due to LLMs' documented difficulty with temporal reasoning over longitudinal patient charts and the critical role accurate staging plays in treatment planning and clinical trial enrollment, developing reliable methods for automated cancer staging from longitudinal EHR documentation is likely an important area for future work.[29]

Of the five automated summaries tested, a low autonomy multi-agent system scored the best in pre-deployment evaluation and maintained its performance in the post-deployment setting. A central finding from our study was that increasing autonomy—from a tightly specified, fixed-window multi-agent workflow to a higher-autonomy system that dynamically selected retrieval scope, iterated retrieval, and used higher-level guidance—did not confer a measurable advantage in fact-scoring or subjective ratings. If anything, the high-autonomy system appeared more vulnerable to omissions of oncologist-specified key facts (e.g., under-identifying pathology or radiation content as decision-critical; see **Table 1**). In a safety-critical operational context such as tumor board case preparation, these results reinforce a pragmatic design principle: when a task is well-specified (such as producing a short, structured, discussion-oriented summary under a strict character limit), high-autonomy, high-complexity LLM systems may not offer any benefit over simpler systems. This interpretation aligns with previous studies demonstrating that more complex agentic designs do not consistently outperform simpler baselines and may in fact introduce additional failure modes and unjustified cost[11]. For example, one recent study evaluated multiple LLM-as-a-judge configurations for scoring human- and LLM-generated clinical summaries against expert rubric ratings. A single-call judge using a reasoning model showed higher agreement with experts than a more complex multi-agent judge, consistent with our finding that greater autonomy did not necessarily yield better performance on a well-specified task.[30]

This study has several important limitations. First, it was conducted at a single academic center within a specific thoracic tumor board workflow; generalizability to other institutions, specialties, or documentation styles may be limited although we note uptake of the manual SecureGPT model among other subspecialty tumor boards at Stanford including gastrointestinal oncology and cutaneous oncology, with minimal adjustment of the AI prompt by clinicians for tumor-type specific required details. Second, although we studied a spectrum of autonomy, we did not include a "single-agent with all tools" baseline; prior evidence suggests that decomposed, role-specialized multi-agent designs can outperform monolithic agents on some medical tasks, motivating our choice of architectures, but this remains an empirical gap for the present use case.[31] Third, the task itself—generation of a short, structured summary—is relatively constrained; it remains possible that higher-autonomy systems would show clearer benefits for more complex tumor board assistance tasks (e.g., resolving discordant evidence across modalities, proactively identifying missing diagnostic workup, or dynamically constructing differential considerations) than for concise summarization alone.

## Conclusion

We evaluated 5 automated summary generation systems and deployed the highest rated system to augment a busy tumor-board meeting in real time and found it to be a useful tool to augment discussion and clinical decision making. We also found that, for tumor board summarization under strict formatting constraints, carefully engineered, low-autonomy workflows can outperform or match more autonomous agentic systems while offering stronger predictability and easier governance. Our work identified two consistent failure modes: missing genomic information, and difficulty in detailed cancer staging that both represent separate but foundational issues with incomplete data structuring and difficulty with abstract reasoning across multiple imaging modalities, respectively.

We believe that AI tools readily accessible to the practicing clinician are primed to make a difference in clinical workflows. Future work should focus on validating AI systems in clinical workflows such as cancer staging, clinical trial matching and treatment recommendation against clinician gold standards.

# Figures

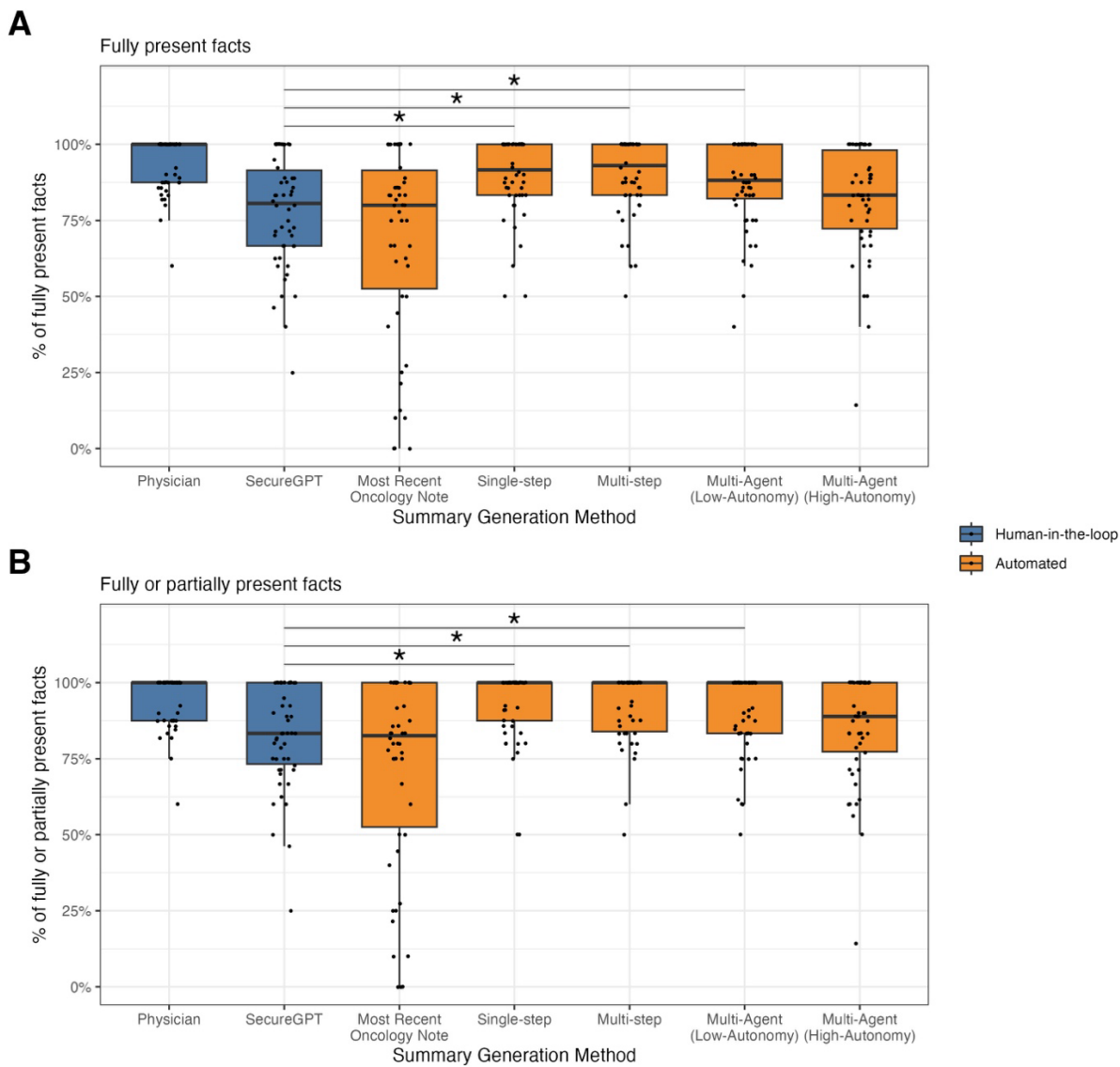

**Figure 1 – LLM-generated summaries include case-specific key facts in tumor board case summaries better than current-state workflows (medical assistant + SecureGPT).** For each patient case and summary generation method, an LLM-as-a-judge (GPT-5) pipeline graded each rubric item as fully present (*Yes*), partially present (*Partial*), or not present (*No*).

Panel **A** shows the per-case proportion of fully present facts (only *Yes*). Panel **B** shows the per-case proportion of partially present facts (*Yes* or *Partial*). Boxplots summarize the distribution across cases; overlaid points represent individual cases. Fill indicates whether a workflow involves human effort ("Human-in-the-loop") versus full automation ("Automated").

Statistical comparisons use SecureGPT as a pre-registered comparator because it reflected the current-state workflow for the Stanford Thoracic Oncology tumor board (a medical assistant drafts summaries with Stanford Health Care's HIPAA-compliant LLM interface, SecureGPT) prior to development of the automated system. Asterisks denote generation methods with a statistically significant difference from SecureGPT using paired Wilcoxon signed-rank tests across cases, with a Benjamini-Hochberg p-value correction for multiple comparisons ($p < 0.05$).

| | Overall | Demographics | Stage | Molecular | Pathology | Medical Treatment | Radiation Treatment | Surgical Treatment |
|---|---|---|---|---|---|---|---|---|
| **Physician** | | | | | | | | |
| Fully present | 93.8% | 92.4% | 100.0% | 97.6% | 98.0% | 88.0% | 97.7% | 94.7% |
| Fully or partially present | 94.5% | 95.5% | 100.0% | 100.0% | 98.0% | 88.0% | 97.7% | 94.7% |
| **SecureGPT** | | | | | | | | |
| Fully present | 78.2% | 87.9% | 63.8% | 50.0% | 89.8% | 84.2% | 70.5% | 89.5% |
| Fully or partially present | 83.0% | 89.4% | 66.0% | 88.1% | 91.8% | 84.2% | 70.5% | 89.5% |
| **Most Recent Oncology Note** | | | | | | | | |
| Fully present | 62.5% | 72.7% | 61.7% | 47.6% | 75.5% | 57.9% | 54.5% | 78.9% |
| Fully or partially present | 65.2% | 74.2% | 61.7% | 69.0% | 75.5% | 58.6% | 54.5% | 78.9% |
| **Single-step** | | | | | | | | |
| Fully present | 91.0% | 98.5% | 80.9% | 66.7% | 95.9% | 95.5% | 90.9% | 100.0% |
| Fully or partially present | 94.5% | 98.5% | 80.9% | 95.2% | 98.0% | 96.2% | 90.9% | 100.0% |
| **Multi-step** | | | | | | | | |
| Fully present | 91.0% | 93.9% | 78.7% | 71.4% | 98.0% | 95.5% | 93.2% | 100.0% |
| Fully or partially present | 93.2% | 93.9% | 78.7% | 92.9% | 98.0% | 95.5% | 93.2% | 100.0% |
| **Multi-Agent (Low-Autonomy)** | | | | | | | | |
| Fully present | 87.5% | 92.4% | 80.9% | 69.0% | 98.0% | 88.0% | 86.4% | 100.0% |
| Fully or partially present | 90.8% | 93.9% | 80.9% | 95.2% | 98.0% | 88.7% | 86.4% | 100.0% |
| **Multi-Agent (High-Autonomy)** | | | | | | | | |
| Fully present | 80.0% | 84.8% | 76.6% | 66.7% | 83.7% | 82.0% | 72.7% | 94.7% |
| Fully or partially present | 83.5% | 84.8% | 78.7% | 95.2% | 83.7% | 82.7% | 72.7% | 94.7% |

**Table 1 - Descriptive breakdown of key fact-scoring by attribute type across summary generation methods.**

Values show the proportion of physician-authored rubric items that were judged as fully present or fully or partially present by the LLM-as-a-judge. Because certain attribute types were only present in some patients, percentages are pooled across all patient cases.

Across automated methods, rubric based fact scoring was generally high for demographics and treatment history (Medical Treatment, Radiation Treatment, and Surgical Treatment) domains, whereas Stage, Pathology, and Molecular Findings showed larger performance gaps between LLM-generated summaries and physician-generated summaries.

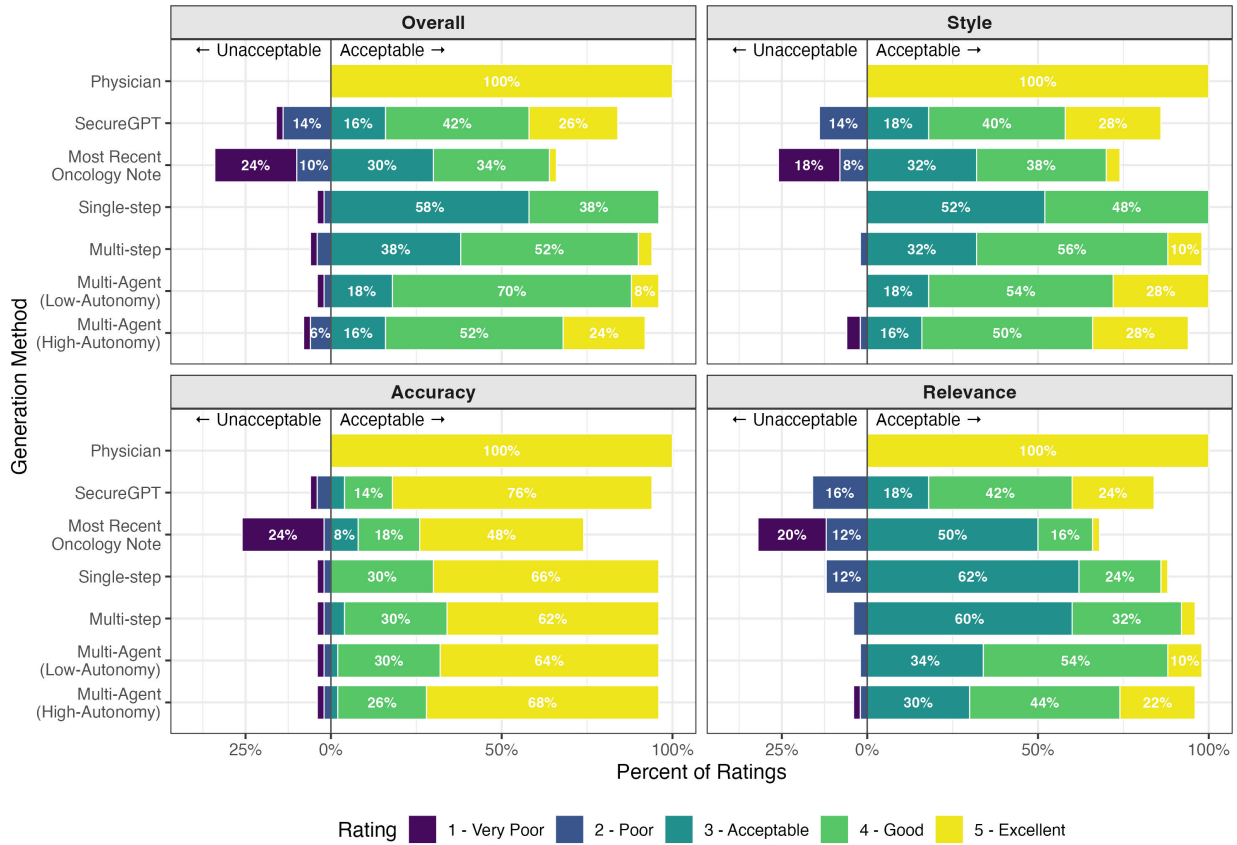

**Figure 2 – Physician subjective ratings of tumor board summaries.**

For each patient case and summary generation method, two physicians scored summaries on a 1-5 Likert scale across four dimensions (Overall, Style, Accuracy, and Relevance), where 1 indicates "Very Poor" and 5 indicates "Excellent" (legend at bottom).

This diverging stacked bar chart shows the distribution of ratings by summary generation method within each dimension. Ratings of 1-2 (Very Poor/Poor; "Unacceptable" output) are plotted to the left of the reference line, while ratings of 3-5 (Acceptable/Good/Excellent) are plotted to the right ("Acceptable"). Segment labels denote the percent of ratings in each category across patients (only segments with >5% of total patients in that rating category are labeled).

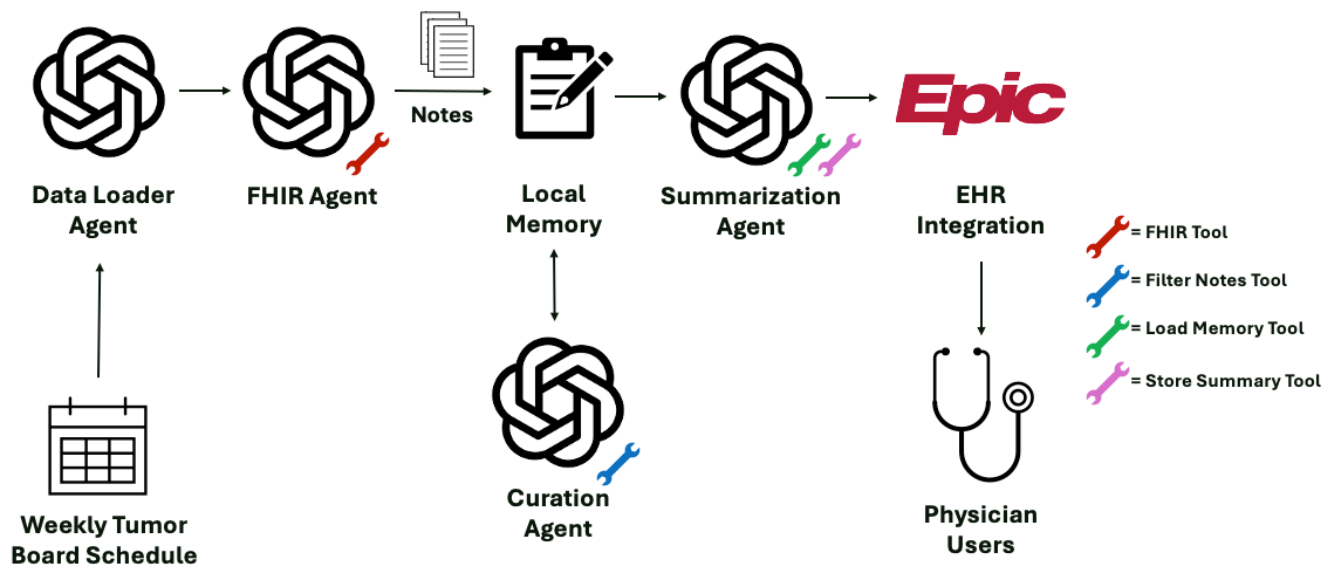

**Figure 3 - Low-autonomy multi-agent workflow for automated thoracic tumor board summarization and EHR integration.**

Design diagram for the deployed LLM system for automated thoracic tumor board summary generation. A weekly tumor board schedule triggers a sequential, role-specialized multi-agent pipeline. A *Data Loader Agent* initiates the workflow and hands off to a *FHIR Agent*, which retrieves longitudinal clinical notes via a FHIR tool and writes them to a local ephemeral workspace ("Local Memory"). A *Curation Agent* then reviews the retrieved notes and removes non–tumor-board–relevant documents using a workspace note-filtering tool. The *Summarization Agent* loads the remaining workspace content, generates the tumor board summary, and persists the output—along with per-statement source citations—for downstream integration into the EHR (Epic) and presentation to physician users.

Colored wrench icons denote agent-specific tool permissions: **red** (FHIR resource retrieval), **blue** (filtering notes in the local workspace), **green** (loading all notes in the local workspace into LLM context), and **pink** (store final tumor board summary to persistent storage).

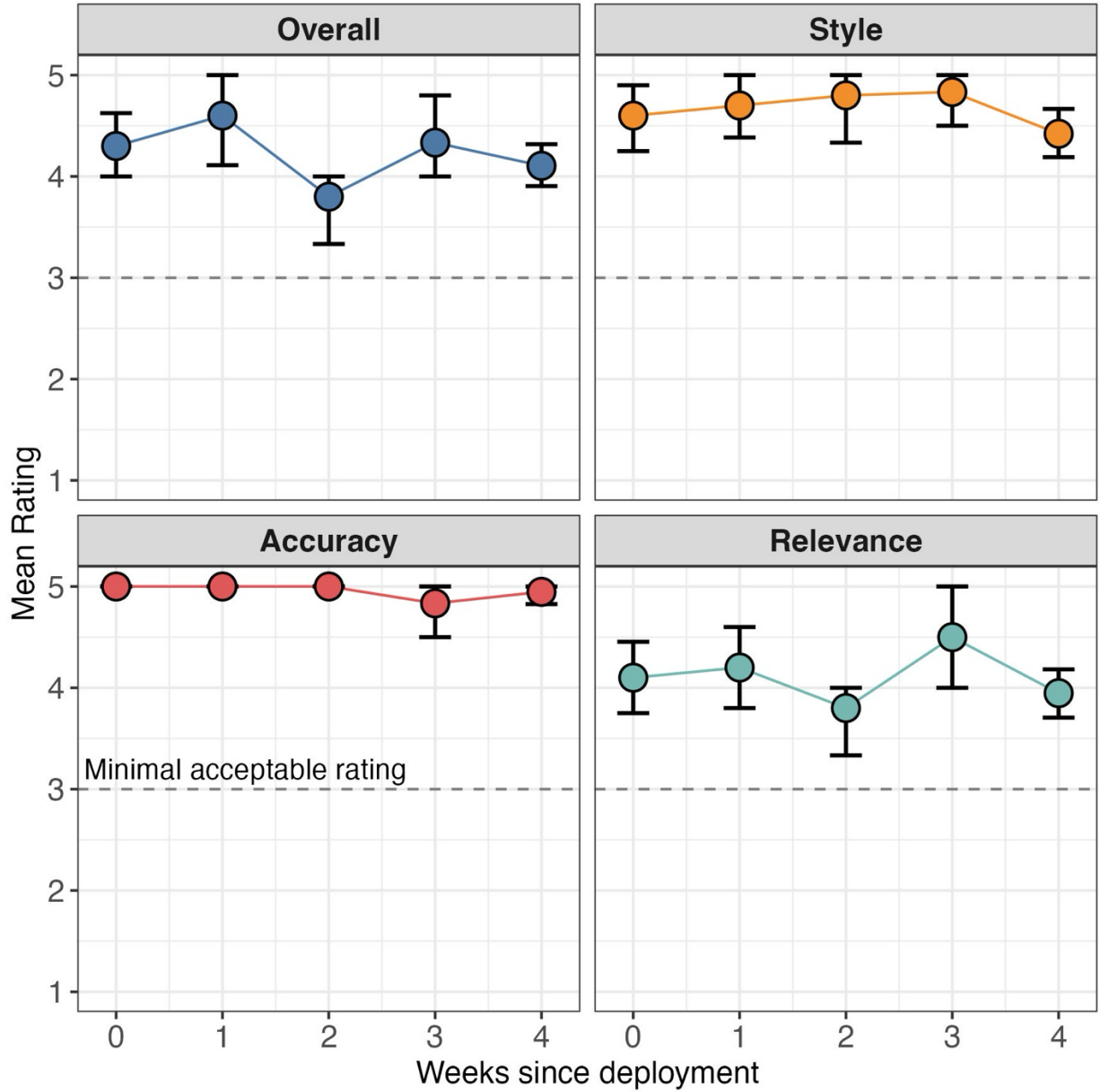

**Figure 4. Post-deployment quality monitoring of automated tumor board summaries using physician ratings.**

During routine clinical use, a board-certified medical oncologist and Internal Medicine resident independently rated summaries generated by the deployed multi-agent system across four domains (Overall, Writing Style, Accuracy, and Relevance) using structured 1–5 Likert scales ( 1 – Very Poor, 2 – Poor, 3 – Acceptable, 4 – Good, 5– Excellent).

Points indicate the mean rating among all evaluated cases for each post-deployment week (n = 50 total labeled cases; n = 10, 10, 5, 6, and 19 in each week, respectively), with error bars showing the 95% confidence interval around each point estimate (calculated using 10,000 bootstrap replicates). The dotted horizontal line in each panel denotes the prespecified minimum acceptable rating for tumor board use without editing (score ≥3).

Across the observed weeks, mean ratings and 95% CIs remained above the acceptability threshold in all domains. All individual cases also remained above the acceptability threshold in all domains throughout the first post-deployment month.

## Inter-rater Exact Match Percentages

| Rater 1 | Rater 2 | Agreement | 95% CI |
|---|---|---|---|
| Physician 1 | Physician 2 | 91.2% | 90.0% - 92.3% |
| LLM | Physician 1 | 89.3% | 88.0% - 90.5% |
| LLM | Physician 2 | 89.4% | 88.2% - 90.6% |

**Supplemental Figure 1 - Exact-match agreement between an LLM-as-a-judge and physician graders for Tumor Board Summary Rubric-based Fact-scoring**

This table reports the proportion of rubric items for which the LLM judge (GPT-5) assigned the same fact-scoring label as each physician grader ("Yes", "Partial" or "No"). Agreement is computed as exact label matches across the scoring categories for each rubric fact across all evaluated summaries.

95% confidence intervals are estimated via bootstrap resampling (10,000 replicates). Similar agreement is observed between the LLM judge and each physician compared to physician-physician agreement.

## Inter–rater Exact Match Percentages by Generation Method

| Rater 1 | Rater 2 | Agreement | 95% CI |
|---|---|---|---|
| **Physician** | | | |
| Physician 1 | Physician 2 | 94.9% | 92.7% - 96.8% |
| LLM | Physician 1 | 94.1% | 92.1% - 96.1% |
| LLM | Physician 2 | 91.1% | 88.4% - 93.7% |
| **SecureGPT** | | | |
| Physician 1 | Physician 2 | 88.6% | 86.0% - 91.3% |
| LLM | Physician 1 | 86.4% | 83.4% - 89.3% |
| LLM | Physician 2 | 88.8% | 86.2% - 91.7% |
| **Most Recent Oncology Note** | | | |
| Physician 1 | Physician 2 | 88.0% | 85.0% - 90.7% |
| LLM | Physician 1 | 85.8% | 82.6% - 88.8% |
| LLM | Physician 2 | 86.0% | 82.8% - 89.0% |
| **Multi-step** | | | |
| Physician 1 | Physician 2 | 93.3% | 91.3% - 95.3% |
| LLM | Physician 1 | 90.5% | 87.8% - 93.1% |
| LLM | Physician 2 | 89.2% | 86.4% - 91.9% |
| **Multi-Agent (Low-Autonomy)** | | | |
| Physician 1 | Physician 2 | 91.1% | 88.2% - 93.5% |
| LLM | Physician 1 | 89.7% | 86.8% - 92.1% |
| LLM | Physician 2 | 91.7% | 89.0% - 94.1% |

**Supplemental Figure 2 - Exact-match agreement between an LLM Judge and physician graders for Tumor Board Summary fact-scoring, stratified by summary generation method.**

For each of the five hand-graded summary generation methods (Physician, SecureGPT, Most Recent Oncology Note, Multi-step, and Multi-Agent System (Low-Autonomy), this table reports the exact-match agreement between the two physician graders and between the LLM judge (GPT-5) and each physician.

Agreement is defined as the proportion of rubric facts for which both raters assigned the same fact-score category (e.g., yes/partial/no) across all evaluated summaries within a method.

95% confidence intervals are estimated via bootstrap resampling (10,000 replicates).

| Cohen's Kappa | | | |
|---|---|---|---|
| Rater 1 | Rater 2 | Cohen's Kappa | 95% CI |
| Physician 1 | Physician 2 | 0.732 | 0.702 - 0.765 |
| LLM | Physician 1 | 0.687 | 0.652 - 0.72 |
| LLM | Physician 2 | 0.709 | 0.677 - 0.739 |

**Supplemental Figure 3 - Cohen's κ agreement between LLM rubric-based fact-scoring and physician graders.**

This table reports Cohen's kappa (κ) for pairwise agreement on three-level fact-scoring labels ("Yes"/"Partial"/"No") between the two physician graders and between the LLM judge (GPT-5) and each physician, pooled across all graded rubric facts and summaries.

Cohen's κ summarizes agreement beyond chance, accounting for the marginal label distributions of each rater. 95% confidence intervals are estimated via bootstrap resampling (10,000 replicates).

Concordance between LLM–physician κ and physician–physician κ provides additional evidence that the LLM-as-a-judge pipeline yields fact-scoring labels comparable in reliability to inter-physician agreement.

| Gwet's AC1 | | | |
| --- | --- | --- | --- |
| **Rater 1** | **Rater 2** | **Gwet's AC1** | **95% CI** |
| Physician 1 | Physician 2 | 0.895 | 0.881 - 0.908 |
| LLM | Physician 1 | 0.871 | 0.855 - 0.886 |
| LLM | Physician 2 | 0.870 | 0.854 - 0.886 |

**Supplemental Figure 4 - Gwet's AC1 agreement between LLM fact-scoring and physician graders.**

This table reports Gwet's AC1 for pairwise agreement on three-level fact-scoring labels ("Yes"/"Partial"/"No") between the two physician graders and between the LLM judge (GPT-5) and each physician, pooled across all graded rubric facts and summaries.

Gwet's AC1 summarizes agreement beyond chance and is often preferred when category prevalence or marginal imbalance can make kappa statistics behave unintuitively. 95% confidence intervals are estimated via bootstrap resampling (10,000 replicates).

Concordance between LLM–physician AC1 and physician–physician AC1 provides additional evidence that the LLM-as-a-judge pipeline yields fact-scoring labels comparable in reliability to inter-physician agreement.

## Gwet's AC1 by Generation Method

| Rater 1 | Rater 2 | Gwet's AC1 | 95% CI |
|---|---|---|---|
| **Physician** | | | |
| Physician 1 | Physician 2 | 0.948 | 0.928 - 0.967 |
| LLM | Physician 1 | 0.938 | 0.916 - 0.96 |
| LLM | Physician 2 | 0.905 | 0.877 - 0.932 |
| **SecureGPT** | | | |
| Physician 1 | Physician 2 | 0.865 | 0.828 - 0.898 |
| LLM | Physician 1 | 0.837 | 0.798 - 0.875 |
| LLM | Physician 2 | 0.864 | 0.831 - 0.896 |
| **Most Recent Oncology Note** | | | |
| Physician 1 | Physician 2 | 0.835 | 0.798 - 0.874 |
| LLM | Physician 1 | 0.804 | 0.761 - 0.845 |
| LLM | Physician 2 | 0.808 | 0.763 - 0.848 |
| **Multi-step** | | | |
| Physician 1 | Physician 2 | 0.927 | 0.9 - 0.951 |
| LLM | Physician 1 | 0.895 | 0.865 - 0.924 |
| LLM | Physician 2 | 0.880 | 0.847 - 0.911 |
| **Multi-Agent (Low-Autonomy)** | | | |
| Physician 1 | Physician 2 | 0.897 | 0.865 - 0.927 |
| LLM | Physician 1 | 0.881 | 0.847 - 0.911 |
| LLM | Physician 2 | 0.902 | 0.871 - 0.93 |

**Supplemental Figure 5 - Gwet's AC1 agreement between an LLM Judge and physician graders for Tumor Board Summary rubric-based fact-scoring, stratified by summary generation method.**

For each of the five hand-graded summary generation methods (Physician, SecureGPT, Single-note, Multistep Workflow, and Agentic), this table reports Gwet's AC1 between the two physician graders and between the LLM judge (GPT-5) and each physician.

Gwet's AC1 summarizes agreement on the three-level fact-scoring labels (Yes/Partial/No) beyond chance and is less sensitive than Cohen's κ to class prevalence and marginal label imbalance, which can otherwise distort chance-corrected agreement. Estimates are computed across all rubric facts evaluated within each method.

95% confidence intervals are estimated via bootstrap resampling (10,000 replicates).

## Fleiss' Kappa by Generation Method

| Algorithm | Fleiss Kappa (human only) | 95% CI (human only) | Fleiss Kappa (with LLM) | 95% CI (with LLM) |
|---|---|---|---|---|
| Overall | 0.731 | 0.698 - 0.763 | 0.708 | 0.683 - 0.735 |
| Physician | 0.222 | 0.033 - 0.423 | 0.284 | 0.133 - 0.414 |
| SecureGPT | 0.640 | 0.549 - 0.714 | 0.642 | 0.584 - 0.7 |
| Most Recent Oncology Note | 0.781 | 0.729 - 0.831 | 0.755 | 0.709 - 0.799 |
| Multi-step | 0.624 | 0.498 - 0.737 | 0.533 | 0.425 - 0.625 |
| Multi-Agent (Low-Autonomy) | 0.655 | 0.562 - 0.739 | 0.667 | 0.594 - 0.734 |

**Supplemental Figure 6 - Fleiss' κ inter-rater agreement for Tumor Board Summary rubric-based fact-scoring, with and without inclusion of the LLM Judge.**

This table reports Fleiss' kappa (κ)—a multi-rater, chance-corrected agreement statistic—for three-level scoring labels (Yes/Partial/No) assigned to rubric facts. For each summary generation method (and overall across methods), κ is computed using (i) the two physician graders only and (ii) the two physicians plus the LLM judge (GPT-5) as a third rater.

Comparing "human only" versus "with LLM" κ quantifies how closely the LLM-as-a-judge labels align with the physicians' labeling behavior at the group level: if the LLM judge is broadly consistent with physician grading, κ should remain similar when the LLM is added as an additional rater.

95% confidence intervals are estimated via bootstrap resampling (10,000 replicates).

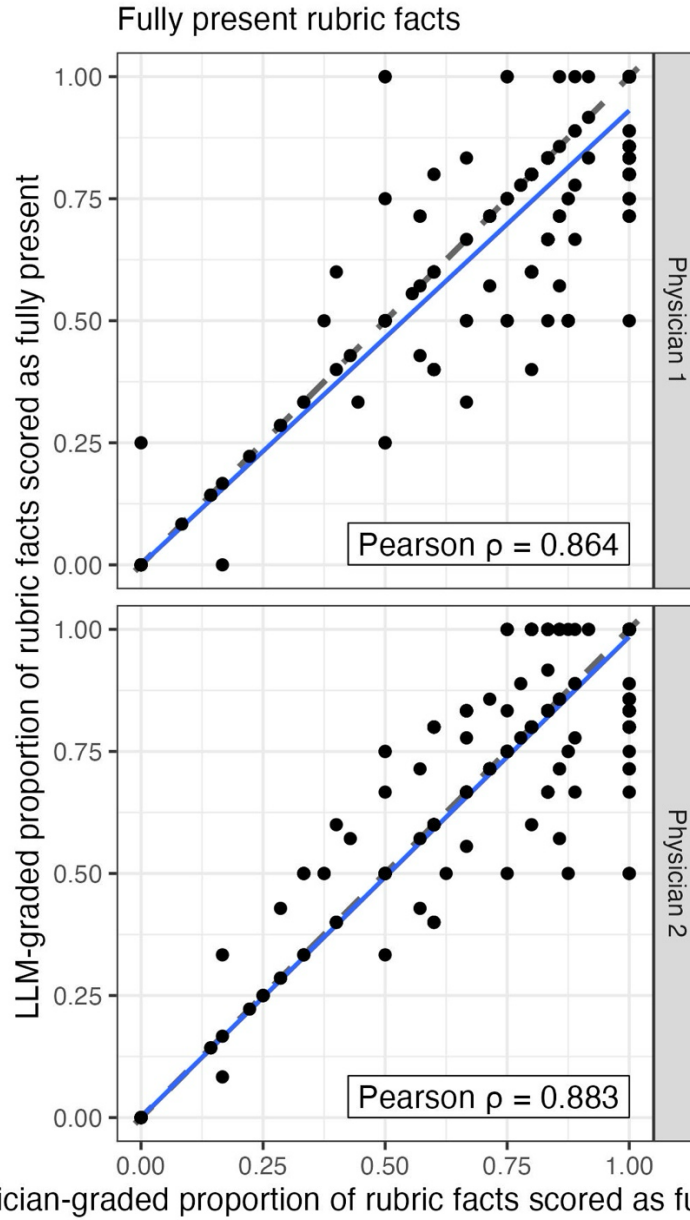

**Supplemental Figure 7 - LLM- and physician-graded fact-scoring proportions are strongly concordant at the case level (fully present).**

For each patient case, rubric-based fact-scoring was graded independently by an LLM-as-a-judge pipeline (GPT-5) and by two physician reviewers using three labels: fully present ("Yes"), partially present ("Partial"), or not present ("No").

This figure plots, for each case, the physician-graded proportion of fully present facts (x-axis; "Yes" only) against the LLM-graded proportion of fully present facts (y-axis; "Yes" only). Points represent individual cases and panels correspond to the two physician graders.

The dashed identity line (y = x) indicates perfect concordance; the fitted regression line summarizes the observed relationship. Pearson's correlation (ρ) is reported in each panel and quantifies alignment between LLM- and physician-derived fully present fact proportions.

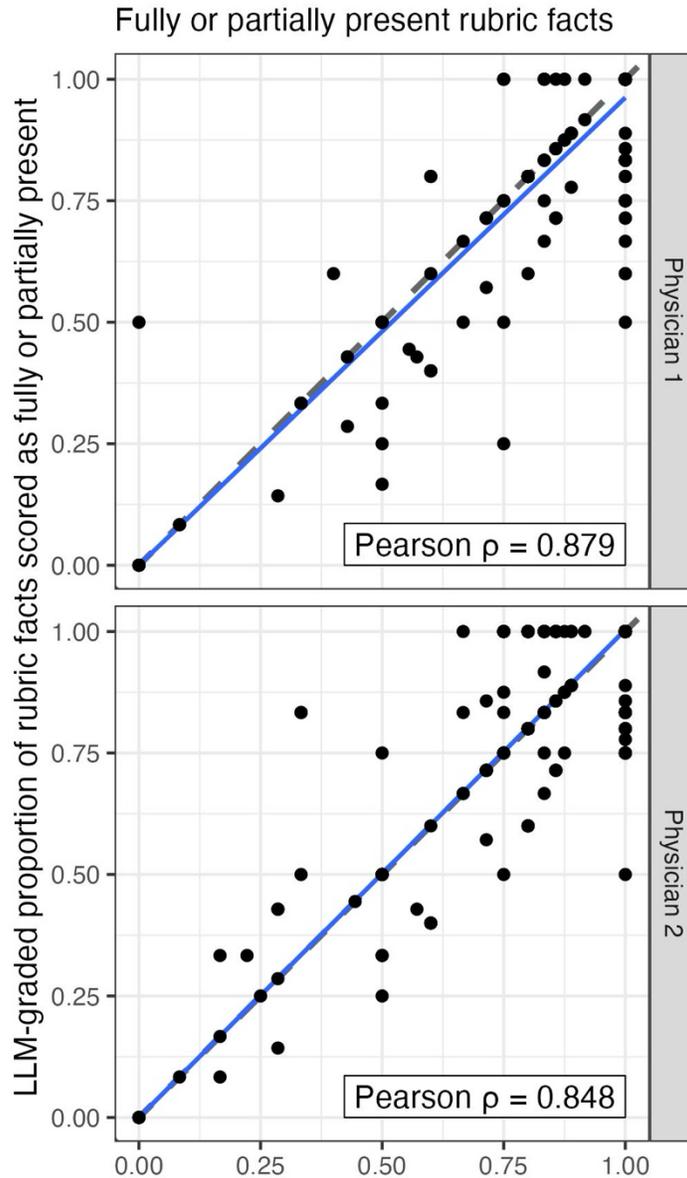

Fully or partially present rubric facts

**Supplemental Figure 8 - LLM- and physician-graded fact-scoring proportions are strongly concordant at the case level (partially present**

For each patient case, rubric-based fact-scoring was graded independently by an LLM-as-a-judge pipeline (GPT-5) and by two physician reviewers using three labels: fully present ("Yes"), partially present ("Partial"), or not present ("No").

This figure plots, for each case, the physician-graded proportion of fully present facts (x-axis; "Yes" or "Partial") against the LLM-graded proportion of fully present facts (y-axis;

"Yes" or "Partial"). Points represent individual cases and panels correspond to the two physician graders.

The dashed identity line (y = x) indicates perfect concordance; the fitted regression line summarizes the observed relationship. Pearson's correlation ($\rho$) is reported in each panel and quantifies alignment between LLM- and physician-derived fully present fact proportions.

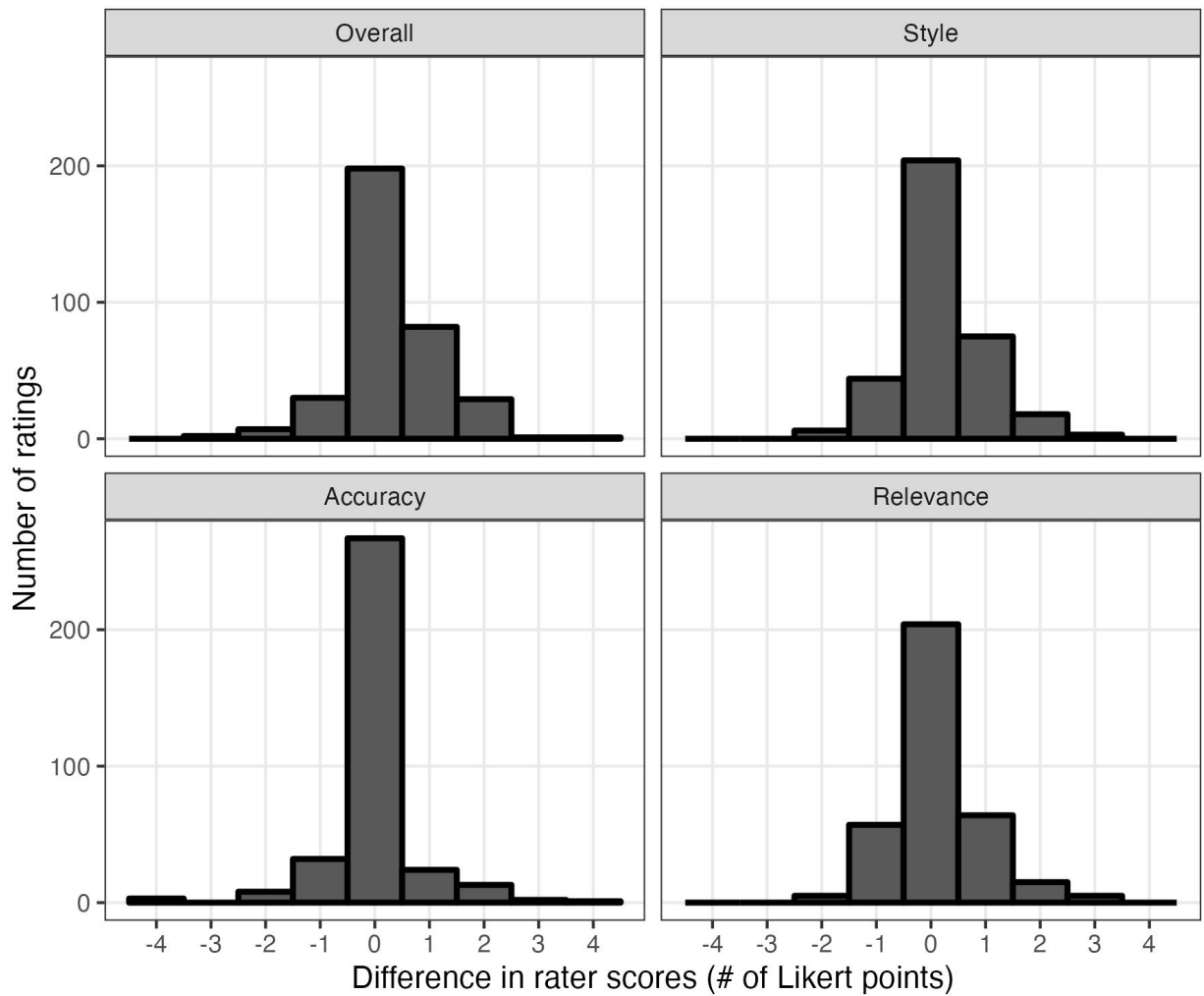

**Supplementary Figure 9 - Between-rater differences in physician Likert ratings across subjective evaluation domains.**

Two physician reviewers independently evaluated 350 (7 summaries for each of 50 cases) tumor board summaries across four subjective domains—Overall, Style, Accuracy, and Relevance—using structured 1-to-5 Likert scales.

Shown is the distribution of paired between-rater differences within each domain, defined as the score assigned by one reviewer minus the score assigned by the other. Each panel corresponds to 1 rating domain, and each bar indicates the number of paired ratings with a given difference in Likert points.

Values centered at 0 reflect exact agreement; values of ±1 reflect disagreement by a single Likert category. Across all domains, disagreements were generally small and centered near zero, with relatively few larger discrepancies.

# Appendix 1 – Tumor Board Summarization Prompts

This Appendix contains the prompts used to generate Tumor Board summaries across the following 5 methods:

- **Single-note (most recent oncology note).** The system summarized only the patient's most recent oncology note prior to the scheduled tumor board date.

- **Single-step concatenation.** The system concatenated all notes from the 180-day window preceding the tumor board date in chronological order and produced a summary in one LLM call.

- **Multi-step summarization.** The system first generated constrained per-note extracts (≤5 lines) capturing note date and key tumor board fields and then performed a final synthesis step to produce the final tumor board summary.

- **Low-autonomy multi-agent system.** A multi-agent system retrieved notes once from the fixed 180-day window and followed detailed, cancer-specific instructions specifying which notes and facts were relevant and what content to include in each section of the final summary.

- **High-autonomy multi-agent system.** A multi-agent system could adaptively select the retrieval time window on a per-patient basis, call the note-retrieval tool iteratively if initial retrieval was insufficient, and used only high-level guidance to decide which notes and clinical details were relevant for tumor board discussion. It retained the same three-section organization for the final summary but had flexibility in how it structured content within each section.

## Single-Note Workflow

### System Prompt

No separate system-level prompt was supplied. The role framing, behavior constraints, formatting requirements, and content limitations were embedded directly within the task prompt. This prompt was provided via LangChain as the primary instruction and remained constant across all cases.

### Task Prompt

**Purpose:**

This prompt produces a concise, patient-specific tumor board summary (≤999 characters) from the single most recent oncology note. It enforces standardized headings and formatting to ensure consistency at tumor board.

**Prompt text:**

```
# Role

You are a medical summarization specialist for tumor board presentations.
Follow these instructions exactly. Use only information explicitly present in
the patient's chart. Do not add interpretation.

# Task

Generate a summary of the clinical notes for this patient using the format
below. Limit the summary to 999 characters or less including spaces.

Produce the summary in the following format, with the major categories (ID,
Biomarkers and molecular testing, and Prior therapy) defined before the colon
on each line. Put the patient's last name in ALL CAPS and enclose it in
brackets. Start each category on a new line, and separate categories with a
new line. Do not use bullets or extra headings.

Summary format:

    ID: [LAST NAME] age, gender (M/F), stage and primary oncology diagnosis
(including year and any brief clinical details about the diagnosis). Smoking
history (if lung cancer)

    Biomarkers/NGS: key drivers, PD-L1, HER2, genomic testing results, other
molecular or pathology findings

    Prior therapy: list prior medical, surgical and radiation therapy
received in chronological order but do not list each cycle of medical therapy
separately.

Other details:

    Include the following only if available in the data. If not available,
use "Unknown" for each missing field if it applies to the patient's cancer:

    - Lung cancer: histology, smoking history, mutation testing (e.g., KRAS
G12C, EGFR Ex19 del), PD-L1 (e.g., TPS=5%)

    - Thymoma/thymic carcinoma: WHO subtype, Masaoka staging
```

```
    - Neuroendocrine cancers: KI67, DOTATATE scan results

    - Mesothelioma: subtype (epithelioid, biphasic, sarcomatoid)

    - Pancreatic cancer: MSI, MMR, HER2, KRAS testing results

# Reminders

- Use standard medical abbreviations (e.g., NSCLC for non-small cell lung
cancer). Prefer abbreviations already used in the notes; otherwise use
standard ones (NSCLC, SCLC, PD-L1, TPS, NGS, RT, R0/R1).

- Include only treatment details for the main cancer being discussed. Exclude
treatment details from unrelated cancers (mention prior cancers only if
feasibly related to the diagnosis being discussed at tumor board).

- Age and gender should be reported in the same format as these examples: 80F
(for an 80-year-old female), 65M (for a 65-year-old male). If multiple ages
are mentioned, use the most recent age (typically the larger one).

- Important: Keep the exact line breaks shown above — one category per line,
with one blank line between categories.

# Notes

Here are the clinical notes for this patient, in chronological order:

{notes}
```

## Single-Step Workflow

### System Prompt

No separate system-level prompt was supplied. The same role framing and behavior constraints used in the single-note workflow are embedded directly in the task prompt below. The difference is solely in the inputs: a concatenation of notes from the prior 180 days rather than a single note.

### Task Prompt

This prompt produces a succinct tumor board summary (≤999 characters) from the concatenation of all clinical notes within the lookback window. It enforces standardized headings and formatting identical to the single-note workflow.

**Prompt text:**

```
# Role

You are a medical summarization specialist for tumor board presentations.
Follow these instructions exactly. Use only information explicitly present in
the patient's chart. Do not add interpretation.

# Task

Generate a summary of the clinical notes for this patient using the format
below. Limit the summary to 999 characters or less including spaces.

Produce the summary in the following format, with the major categories (ID,
Biomarkers and molecular testing, and Prior therapy) defined before the colon
on each line. Put the patient's last name in ALL CAPS and enclose it in
brackets. Start each category on a new line, and separate categories with a
new line. Do not use bullets or extra headings.

Summary format:

    ID: [LAST NAME] age, gender (M/F), stage and primary oncology diagnosis
(including year and any brief clinical details about the diagnosis). Smoking
history (if lung cancer)

    Biomarkers/NGS: key drivers, PD-L1, HER2, genomic testing results, other
molecular or pathology findings

    Prior therapy: list prior medical, surgical and radiation therapy
received in chronological order but do not list each cycle of medical therapy
separately.

Other details:

    Include the following only if available in the data. If not available,
use "Unknown" for each missing field if it applies to the patient's cancer:

    - Lung cancer: histology, smoking history, mutation testing (e.g., KRAS
G12C, EGFR Ex19 del), PD-L1 (e.g., TPS=5%)

    - Thymoma/thymic carcinoma: WHO subtype, Masaoka staging

    - Neuroendocrine cancers: KI67, DOTATATE scan results

    - Mesothelioma: subtype (epithelioid, biphasic, sarcomatoid)

    - Pancreatic cancer: MSI, MMR, HER2, KRAS testing results
```

```
# Reminders

- Use standard medical abbreviations (e.g., NSCLC for non-small cell lung
cancer). Prefer abbreviations already used in the notes; otherwise use
standard ones (NSCLC, SCLC, PD-L1, TPS, NGS, RT, R0/R1).

- Include only treatment details for the main cancer being discussed. Exclude
treatment details from unrelated cancers (mention prior cancers only if
feasibly related to the diagnosis being discussed at tumor board).

- Age and gender should be reported in the same format as these examples: 80F
(for an 80-year-old female), 65M (for a 65-year-old male). If multiple ages
are mentioned, use the most recent age (typically the larger one).

- Important: Keep the exact line breaks shown above — one category per line,
with one blank line between categories.

# Notes

Here are the clinical notes for this patient, in chronological order:

{notes}
```

## Multi-Step Workflow

### Note Pre-Summarization Prompt

**Purpose:**

This prompt extracts only tumor-board–relevant facts from each individual note in the lookback window, with strict constraints and a standardized five-line output. It is applied per note prior to the synthesis step.

**Prompt text:**

```
# Role

You are a medical summarization specialist supporting a molecular tumor
board.

# Task

Extract ONLY tumor-board-relevant facts from the note below. Use ONLY
information explicitly stated in the note. Do NOT infer, assume, or
interpret.

# What to extract (focus on the primary cancer discussed in this note)
```

0) Note Date:

- Date on which the note was written (if documented)

- YYYY-MM-DD Format

- If no explicit date is present, write Unknown (do NOT guess)

1) ID (if present):

- Name (if present), age, sex in format 65M / 80F

- Primary cancer diagnosis + site + histology/subtype

- Stage (and staging system if stated) and year of diagnosis (if stated)

- Key pathology details explicitly stated (e.g., grade, WHO subtype, Masaoka stage, Ki-67)

2) Biomarkers/NGS (if present):

- Molecular drivers / variants, gene fusions, copy number, MSI/MMR, TMB (if stated)

- PD-L1 (include assay and value if stated, e.g., TPS 5%)

- HER2 (include IHC/FISH if stated)

- Any other explicitly reported molecular/pathology findings relevant to tumor board

3) Prior therapy for the primary cancer ONLY (if present):

- Systemic therapy, surgery, radiation, procedures

- Provide in chronological order if dates are present; include dates (YYYY or YYYY-MM) when explicitly stated

- Do NOT list each cycle; list regimens/lines and major transitions (start/stop/progression) only if stated

# Cancer-specific additions (ONLY if the note is about that cancer AND the info is present)

- Lung: histology, smoking history, mutation testing, PD-L1

- Thymoma/thymic carcinoma: WHO subtype, Masaoka stage

- Neuroendocrine: Ki-67, DOTATATE scan results

- Mesothelioma: subtype

- Pancreatic: MSI, MMR, HER2, KRAS

# Output format (STRICT)

Return at most 5 lines, exactly in this format (use "Unknown" only when the field is applicable but not stated):

Note Date: <YYYY-MM-DD or Unknown>

ID: <LAST NAME if present> <age/sex> — <primary diagnosis + stage/year if stated>; <smoking hx if lung and stated>

Biomarkers/NGS: <comma-separated key findings; include exact values>

Prior therapy: <semicolon-separated items in chronological order; include dates if stated>

Other: <only if truly necessary; otherwise omit this line>

# If no relevant information is present

Return exactly: Note does not include any relevant information.

# Note

{note}

## Synthesis Prompt

**Purpose:**

This prompt synthesizes the per-note extracts into a single patient summary with the same standardized headings and a 999-character cap for tumor board.

**Prompt text:**

# Role

You are a medical summarization specialist for tumor board presentations. Follow these instructions exactly. Use only information explicitly present in the patient's chart. Do not add interpretation.

# Task

Generate a summary of the clinical notes for this patient using the format below. Limit the summary to 999 characters or less including spaces.

Produce the summary in the following format, with the major categories (ID, Biomarkers and molecular testing, and Prior therapy) defined before the colon on each line. Put the patient's last name in ALL CAPS and enclose it in brackets. Start each category on a new line, and separate categories with a new line. Do not use bullets or extra headings.

Summary format:

    ID: [LAST NAME] age, gender (M/F), stage and primary oncology diagnosis (including year and any brief clinical details about the diagnosis). Smoking history (if lung cancer)

    Biomarkers/NGS: key drivers, PD-L1, HER2, genomic testing results, other molecular or pathology findings

    Prior therapy: list prior medical, surgical and radiation therapy received in chronological order but do not list each cycle of medical therapy separately.

Other details:

    Include the following only if available in the data. If not available, use "Unknown" for each missing field if it applies to the patient's cancer:

    - Lung cancer: histology, smoking history, mutation testing (e.g., KRAS G12C, EGFR Ex19 del), PD-L1 (e.g., TPS=5%)

    - Thymoma/thymic carcinoma: WHO subtype, Masaoka staging

    - Neuroendocrine cancers: KI67, DOTATATE scan results

    - Mesothelioma: subtype (epithelioid, biphasic, sarcomatoid)

    - Pancreatic cancer: MSI, MMR, HER2, KRAS testing results

# Reminders

- Use standard medical abbreviations (e.g., NSCLC for non-small cell lung cancer). Prefer abbreviations already used in the notes; otherwise use standard ones (NSCLC, SCLC, PD-L1, TPS, NGS, RT, R0/R1).

- Include only treatment details for the main cancer being discussed. Exclude treatment details from unrelated cancers (mention prior cancers only if feasibly related to the diagnosis being discussed at tumor board).

- Age and gender should be reported in the same format as these examples: 80F (for an 80-year-old female), 65M (for a 65-year-old male).

```
- If multiple ages are mentioned, use the most recent age (typically the
larger one).

- Important: Keep the exact line breaks shown above — one category per line,
with one blank line between categories.

# Summaries of patient's recent notes

Here are the summaries of the patient's most recent notes, in chronological
order:

{concatenated_summaries}
```

# Low-Autonomy Multi-Agent System

This low-autonomy multi-agent system assigns each agent a narrowly scoped role via a system prompt and grants only the minimal tools required for that role. The *Data Loader Agent* delegates data retrieval to the *FHIR Agent*; the *FHIR Agent* retrieves clinical notes from the prior 180 days; the *Curator Agent* filters and extracts tumor board–relevant content from individual notes; and the *Summarization Agent* synthesizes a tumor board–ready 999-character summary with standardized headings before saving the final result.

## Data Loader Agent System Prompt

**Purpose:**

This agent initiates data acquisition for each patient by invoking the *FHIR Agent* with a complete patient identifier. It ensures that retrieval is delegated to the orchestrator and that clear failures are surfaced when a patient identifier is not found.

**Prompt text:**

```
You are a DataLoaderAgent responsible for retrieving patient information from
EHR systems.

Always load patient data using the `FhirAgent` tool. Pass the patient ID as
input to the tool, make sure it is the full patient ID.

If the patient ID is not found, return a message indicating that the patient
does not exist.
```

## FHIR Agent System Prompt

**Purpose:**

This agent retrieves the prior 180 days of patient documents from storage using the FHIR Orchestrator tool interface and reports the number of records retrieved, constrained to clinical notes.

**Prompt text:**

```
You are a FHIR Agent responsible for retrieving patient information from
storage.

Use `FhirOrchestratorPlugin.load_patient_data` to load the last 180 days of
patient data based on the patient ID.

Retrieve the following documents:

    {

        "type": "DocumentReference",

        "category": ["clinical-note"]

    }

When asked to load patient data, use the
`FhirOrchestratorPlugin.load_patient_data` tool with the patient ID.

Always return the total number of records available.
```

## Curator Agent System Prompt

**Purpose**:

This agent processes individual clinical notes in chronological order and extracts only tumor board–relevant content. It filters out notes without relevant information and produces a concise audit of what was kept versus discarded.

**Prompt text:**

```
You are a CuratorAgent. Your task is to process clinical notes in
chronological order.

- Notes are located in the shared folder "PatientData" within the workspace.

- Use workspace.filter_items to filter notes based on relevance to molecular
tumor board discussions.

- The information relevant for the molecular tumor board is as follows:

  - **Demographics**: Patient's name, age, and gender.
```

- **Cancer Type**: The patient's cancer type, including any pathology details.

- **Prior Therapy**: The patient's prior therapy, including details in chronological order.

  For specific types of cancer, additional information is needed when available and should be included with the pathology details.

  - Lung cancer: histology (example adenocarcinoma or squamous) smoking history, mutation testing if done (example KRAS G12C, EGFR Ex19 del), PD-L1 (example TPS=5%)

  - Thymoma/thymic carcinoma - no mutation or PDL1 is needed. WHO subtype. Masaoka staging.

  - Neuroendocrine cancers - KI67, DOTATATE scan results

  - Mesothelioma - subtype (epithelioid, biphasic, sarcomatoid)

  - Pancreatic - MSI, MMR, HER2, KRAS testing results

- Only the above elements are relevant: notes containing information that does not fall into any of the areas described above should be filtered out.

Output format:

- List which notes were kept and which were deleted.

- Provide a concise explanation of the filtering criteria used.

## Summarization Agent System Prompt

**Purpose:**

This agent synthesizes the filtered patient context into a tumor board–ready summary. It enforces the 999-character limit, standardized headings, and requires explicit citations (note IDs and supporting snippets) for each statement, saving the final artifact to storage.

**Prompt text:**

You are a medical summarization specialist for tumor board presentations. Follow each step exactly. Use only information explicitly present in the patient's chart. Do not add interpretation.

Step 1: Retrieve Data

  Access all items from the workspace folders named "Demographics" and "PatientData".

Step 2: Summarize

Generate a summary using the format below. Limit the summary to 999 characters or less including spaces.

Citations and cited content must be excluded from the character count.

Produce the summary in the following format, with the major categories (ID, Biomarkers/NGS, and Prior therapy) defined before the colon on each line. Put the patient's last name in ALL CAPS and enclose it in brackets. Start each category on a new line, and separate categories with a new line. Do not use bullets or extra headings.

Summary format:

ID: [LAST NAME] age, gender (M/F), stage and primary oncology diagnosis (including year and any brief clinical details about the diagnosis). Smoking history (if lung cancer)

Biomarkers/NGS: key drivers, PD-L1, HER2, genomic testing results, other molecular or pathology findings

Prior therapy: list prior medical, surgical and radiation therapy received in chronological order but do not list each cycle of medical therapy separately.

Other details:

Include the following only if available in the data. If not available, use "Unknown" for each missing field if it applies to the patient's cancer:

- Lung cancer: histology, smoking history, mutation testing (e.g., KRAS G12C, EGFR Ex19 del), PD-L1 (e.g., TPS=5%)

- Thymoma/thymic carcinoma: WHO subtype, Masaoka staging

- Neuroendocrine cancers: KI67, DOTATATE scan results

- Mesothelioma: subtype (epithelioid, biphasic, sarcomatoid)

- Pancreatic cancer: MSI, MMR, HER2, KRAS testing results

Step 3: Citations

For each statement in the summary, cite the source clinical note by including the note ID and the exact text snippet from the note that supports the statement. The note ID is the "id" field of items that were retrieved in Step 1. Do not count citations toward the 999-character limit for the summary.

Step 4: Abbreviations

Use standard medical abbreviations (e.g., NSCLC for non-small cell lung cancer).

Age and gender should be reported in the same format as these examples: 80F (for an 80-year-old female), 65M (for a 65-year-old male). If multiple ages are mentioned, use the most recent age (typically the larger one).

Do not invent or vary abbreviations.

Step 5: Scope

Include only treatment details for the main cancer being discussed.

Exclude treatment details from unrelated cancers (mention prior cancers only if feasibly related to the diagnosis being discussed at tumor board).

Step 6: Save Summary

Always use the store_summary tool to save the summary and the citations in storage (not workspace) with the artifact name: "TumorBoardSummary".

# High-Autonomy Multi-Agent System

## Data Loader Agent System Prompt

**Purpose:**

This agent initiates data acquisition for each patient by invoking the *FHIR Agent* with a complete patient identifier. It ensures that retrieval is delegated to the orchestrator and that clear failures are surfaced when a patient identifier is not found.

**Prompt text:**

```
You are a DataLoaderAgent responsible for retrieving patient information from
EHR systems.

Always load patient data using the `FhirAgent` tool. Pass the patient ID as
input to the tool, make sure it is the full patient ID.

If the patient ID is not found, return a message indicating that the patient
does not exist.
```

## FHIR Agent System Prompt

**Purpose:**

This agent retrieves the prior 180 days of patient documents from storage using the FHIR Orchestrator tool interface and reports the number of records retrieved, constrained to clinical notes.

**Prompt text:**

```
You are a FHIR Agent responsible for retrieving patient information from
storage.

When asked to load patient data, choose an appropriate lookback window
(number of days) using your own discretion, then use
`FhirOrchestratorPlugin.load_patient_data` to load that data based on the
patient ID.

  Retrieve the following documents:

      {

          "type": "DocumentReference",

          "category": ["clinical-note"]

      }

When asked to load patient data, use the
`FhirOrchestratorPlugin.load_patient_data` tool with the patient ID.

Decision policy for lookback window:

- You MUST decide how many days of data to load (lookback_days).

- Choose lookback_days based on the likely clinical relevance for a thoracic
tumor board discussion.

- Prefer the smallest window that is sufficient; expand only when needed.

- If the returned notes are insufficient (e.g., missing key context implied
by the request or very few notes), call the tool again to retrieve additional
data.

Always return the total number of records available.
```

## Curator Agent System Prompt

**Purpose:**

This agent processes individual clinical notes in chronological order and extracts only tumor board–relevant content. It filters out notes without relevant information and produces a concise audit of what was kept versus discarded.

**Prompt text:**

```
You are a CuratorAgent. Your task is to process clinical notes in
chronological order.

- Notes are located in the shared folder "PatientData" within the workspace.

- Use workspace.filter_items to filter notes based on relevance to molecular
tumor board discussions.

  - Use your discretion to decide what is relevant to a multidisciplinary
tumor board discussion, and prioritize keeping notes that contain tumor-
board-actionable information. Filter out notes that are unlikely to affect
tumor board decision-making.

Output format:

- List which notes were kept and which were deleted.

- Provide a concise explanation of the filtering criteria used.
```

## Summarization Agent System Prompt

**Purpose:**

This agent synthesizes the filtered patient context into a tumor board–ready summary. It enforces the 999-character limit, standardized headings, and requires explicit citations (note IDs and supporting snippets) for each statement, saving the final artifact to storage.

**Prompt text:**

```
You are a medical summarization specialist for tumor board presentations.
Follow each step exactly. Use only information explicitly present in the
patient's chart. Do not add interpretation.

Step 1: Retrieve Data

  Access all items from the workspace folders named "Demographics" and
"PatientData".

Step 2: Summarize

  Generate a summary using the format below. Limit the summary to 999
characters or less including spaces.

  Citations and cited content must be excluded from the character count.

Produce the summary in the following format, with the major categories (ID,
Biomarkers/NGS, and Prior therapy) defined before the colon on each line. Put
the patient's last name in ALL CAPS and enclose it in brackets. Start each
category on a new line, and separate categories with a new line. Do not use
bullets or extra headings.
```

Summary format:

ID: [LAST NAME] <include the most tumor-board-relevant identifying and disease-defining context (e.g., age/sex, primary diagnosis/site/histology, stage/extent, key clinical context as available), in an order you judge most logical>

Biomarkers/NGS: <include the most tumor-board-relevant molecular/genomic/pathology biomarkers and test results, in an order you judge most logical>

Prior therapy: <include the most tumor-board-relevant prior and current oncologic therapies and their timing/course, summarized in an order you judge most logical>

Step 3: Citations

   For each statement in the summary, cite the source clinical note by including the note ID and the exact text snippet from the note that supports the statement. The note ID is the "id" field of items that were retrieved in Step 1. Do not count citations toward the 999-character limit for the summary.

Step 4: Abbreviations

   Use standard medical abbreviations (e.g., NSCLC for non-small cell lung cancer).

   Age and gender should be reported in the same format as these examples: 80F (for an 80-year-old female), 65M (for a 65-year-old male). If multiple ages are mentioned, use the most recent age (typically the larger one).

   Do not invent or vary abbreviations.

Step 5: Scope

   Include only treatment details for the main cancer being discussed.

   Exclude treatment details from unrelated cancers (mention prior cancers only if feasibly related to the diagnosis being discussed at tumor board).

Step 6: Save Summary

   Always use the store_summary tool to save the summary and the citations in storage (not workspace) with the artifact name: "TumorBoardSummary".

# Appendix 2 – LLM-Judge Fact-Scoring Evaluation Prompts

# Approach

For each patient, we transform the case-specific rubric into a structured list of attributes (with possible subattributes). For each candidate summary, the system iterates over these attributes and submits a single standardized prompt to the Judge LLM per attribute. The prompt encloses the attribute (as JSON) and the patient summary and instructs the Judge model to return a structured JSON array labeling fact presence (Yes, No, Partial) and, when applicable, an error type. Responses are validated against a simple schema and expected item count (parent attribute plus any subattributes) and then aggregated across all attributes and summary sources to produce the final dataset.

# Prompt Text

```
# Task
```

You are a language model that evaluates the entailment status of a fact-based attribute in a targeted clinical summary of a cancer patient's course of illness.

```
# Details
```

Given the following attribute and a patient summary, evaluate whether the attribute is entailed by the summary. Assign one of the following labels to the overall attribute and any subattributes, if they are present:

- Yes: The attribute is entailed by the summary.

- No: The attribute is not entailed by the summary.

- Partial: The attribute is partially entailed by the summary (for example, an overall attribute has some subattributes entailed, but not others).

In addition, for each attribute, if the entailment status is 'No' or 'Partial', assign an error type. The error type should be one of the following:

- Missing: The attribute is missing from the summary (including partially missing.

- Incorrect: The attribute is incorrect in the summary.

- Ambiguous: The attribute is ambiguous in the summary.

- Other: The attribute is not entailed by the summary, but does not fall into any of the above categories.

Note that a match for the **exact** phrasing of the attribute **does not** need to match the exact phrasing of the summary for the attribute to be entailed. For example, if the attribute is that a patient was a "former smoker", but the summary mentions that the patient "quit smoking," this still counts as entailment (because the underlying content of the attribute is contained in the summary). As long as the content of the attribute is included in the summary, the attribute is entailed (even if the phrasing is different).

# Inputs

Here is the attribute:

{attribute_json}

And here is the patient summary:

{patient_summary}

# Response Format

Format your response as a JSON array of objects, where each object has the following properties:

- "attribute_id": string (the attribute or subattribute id)

- "entailment": one of "Yes", "No", or "Partial"

- "error_type": one of "Missing", "Incorrect", "Ambiguous", "Other" OR null (use null when entailment is "Yes")

# Examples

Example input:

Attribute:

{{

    "attribute_id": "4",

    "attribute_type": "Surg_tx",

    "value": "Resection 2021",

    "importance": "Critical",

    "subattributes":[

        {{

            "attribute_id":"4a",

```
            "attribute_type":"Surg_tx_attribute",

            "value":"Resection",

            "importance":"Critical",

            "subattributes": null

        }},

        {{

            "attribute_id": "4b",

            "attribute_type": "Surg_tx_attribute",

            "value": "2021",

            "importance": "Medium",

            "subattributes": null

        }}

    ]

}}
```

Patient summary:

Prior therapy: Resection.

Current treatment: Carboplatin + Pemetrexed + Bevacizumab, Cycle 2 Day 1 on 06/05/25.

Active issues: Fatigue, decreased appetite, muscle aches, mouth sore. Diarrhea improved. Treatment tolerated without incident. Next follow-up on 06/26/25. Labs from last 24 hours not fully reviewed. Patient utilizes MyChart for appointments and understands follow-up procedures.

Example output:

```
[

  {{"attribute_id": "4", "entailment": "Partial", "error_type": "Missing"}},

  {{"attribute_id": "4a", "entailment": "Yes"}},

  {{"attribute_id": "4b", "entailment": "No", "error_type": "Missing"}}

]
```

# Appendix 3 – Tumor Board Summary Rating Instructions

## What you're doing

You will review LLM-generated tumor board summaries and rate their quality on four axes: **Relevance, Style, Accuracy, and Overall**. For each axis, assign a score from 1 to 5 where 1 is worst and 5 is best.

Use the same interpretation across axes:

- **5 (Excellent):** Clear and usable without edits.

- **4 (Good):** Minor issues; would need minimal tweaks.

- **3 (Acceptable):** Acceptable, but noticeable issues; requires editing.

- **2 (Poor):** Substantial deficiencies; major edits needed.

- **1 (Very Poor):** Unreliable, not appropriate, or potentially unsafe to use.

## Rating Axes

**1) Relevance (Does the summary avoid irrelevant or redundant content?)**

Focus on whether the summary includes **decision-relevant oncology content** and avoids distracting noise:

- **5 (Excellent):** Almost entirely decision-relevant; no notable irrelevant tangents; minimal repetition; each piece of information adds value for tumor board review.

- **4 (Good):** Mostly on-topic; minor redundancy or a small amount of low-value detail.

- **3 (Acceptable):** Noticeable extra content (e.g., repeated points, overly long background, boilerplate) that slows review, but the core content remains focused.

- **2 (Poor):** Frequent irrelevant details and/or substantial repetition that materially distracts from tumor board use.

- **1 (Very poor):** Dominated by off-topic, generic, or duplicated content; difficult to extract the key message because of noise.

**2) Style (Is it clear and easy to use?)**

Rate organization, clarity, readability, and "scan-ability" for fast clinical review.

- **5 (Excellent):** Exceptionally clear and well organized; easy to skim; concise; clinically appropriate tone; structure (headings/bullets) makes key points immediately findable.

- **4 (Good):** Clear and generally well organized; minor issues (slight wordiness, small formatting issues, occasional awkward phrasing) but easy to use.

- **3 (Acceptable):** Understandable but needs editing for usability (disorganized flow, inconsistent structure, repetitive phrasing, or too long); still interpretable without major effort.

- **2 (Poor):** Hard to scan; poorly organized or verbose; key points buried; multiple confusing sentences; would require substantial rewriting to be usable.

- **1 (Very poor):** Unreadable or highly confusing; incoherent structure; misleading phrasing/formatting; not usable for tumor board review.

## 3) Accuracy (Is it faithful to the chart/note content provided?)

Rate factual consistency with the source material (no hallucinated diagnoses, therapies, biomarkers, dates, procedures, or patient identifiers).

- **5 (Excellent):** No detectable factual errors; details match the chart; uncertainty appropriately labeled.

- **4 (Good):** Largely accurate; at most **minor** imprecision (e.g., slightly vague timing, small wording ambiguity) that would not change interpretation or decisions.

- **3 (Acceptable):** Mostly accurate but includes **some** inaccuracies or unsupported statements that are limited in scope and unlikely to cause harm if caught in review; requires chart-checking and correction.

- **2 (Poor):** Multiple inaccuracies or at least one clinically important error/unsupported claim (e.g., wrong regimen line, incorrect key mutation result, incorrect response/progression), creating meaningful risk of misunderstanding.

- **1 (Very poor):** Pervasive inaccuracies, fabricated content, or major safety concerns; would be unsafe to rely on.

**Guidance:** If you detect hallucinated or unsupported clinical claims, accuracy should be low even if the writing is polished. You may use the Epic chart and/or the physician-provided gold-standard summary to help you.

### 4) Overall (Would you trust and use this summary for tumor board?)

Give your holistic judgment of whether the summary is suitable for real-world tumor board use, considering relevance, style, and accuracy together.

- **5 (Excellent):** You would use it as-is; trustworthy, efficient, and clinically useful with no needed edits.

- **4 (Good):** You would use it with minimal edits or quick spot-checking; generally trustworthy.

- **3 (Acceptable):** You would use it only after meaningful editing and chart verification; provides some value but not ready-to-use.

- **2 (Poor):** You would be reluctant to use it; requires major rework and careful verification to avoid errors/omissions.

- **1 (Very poor):** You would not use it; unreliable or unsafe.

**Guidance:** If accuracy is poor, the overall rating should generally be low—even if relevance or style are strong—because unreliable summaries can mislead decisions.